\documentclass{article}

\usepackage{arxiv}

\usepackage[utf8]{inputenc} 
\usepackage[T1]{fontenc}    
\usepackage{hyperref}       
\usepackage{url}            
\usepackage{booktabs}       
\usepackage{amsfonts}       
\usepackage{nicefrac}       
\usepackage{microtype}      
\usepackage{lipsum}		
\usepackage{graphicx}       
\usepackage{subcaption}
\usepackage{array}
\usepackage{multirow}
\usepackage{caption}

\title{Unsupervised Transfer Learning via BERT Neuron Selection}

\author{
  Mehrdad Valipour, En-Shiun Annie Lee, Jaime R.~Jamacaro, Carolina Bessega\\
  Stradigi AI\\
  Montreal, QC, Canada\\
  \texttt{\{mehrdadv, anniee, jaimec, carolinab\}@stradigi.ai} \\
}

\newcommand{\ourProcess}[1][variable_value]{#1TSNS}

\begin{document}
\maketitle

\begin{abstract}
Recent advancements in language representation models such as BERT have led to a rapid improvement in numerous natural language processing tasks. However, language models usually consist of a few hundred million trainable parameters with embedding space distributed across multiple layers, thus making them challenging to be fine-tuned for a specific task or to be transferred to a new domain. To determine whether there are task-specific neurons that can be exploited for unsupervised transfer learning, we introduce a method for selecting the most important neurons to solve a specific classification task. This algorithm is further extended to multi-source transfer learning by computing the importance of neurons for several single-source transfer learning scenarios between different subsets of data sources. Besides, a task-specific fingerprint for each data source is obtained based on the percentage of the selected neurons in each layer. We perform extensive experiments in unsupervised transfer learning for sentiment analysis, natural language inference and sentence similarity, and compare our results with the existing literature and baselines. Significantly, we found that the source and target data sources with higher degrees of similarity between their task-specific fingerprints demonstrate a better transferability property. We conclude that our method can lead to better performance using just a few hundred task-specific and interpretable neurons.

\end{abstract}

\section{Introduction}
It is important to have high-quality training data to build a reliable deep learning model; however, in practice well-labelled data is rarely available for the domain or task of interest.  
Therefore, transfer learning from a resource-rich source domain to a resource-poor target domain is a central focus in natural language processing (NLP) due to the divergent corpus.  

Recent developments in neural language transformers (such as BERT \cite{devlin19bert}, which stands for Bidirectional Encoder Representations from Transformers, ElMo \cite{peters-etal-2018-deep}, and GPT \cite{bottou2014machine}) trained on large text corpora (such as Wikipedia and books) led to groundbreaking linguistic representation.  Despite their novelty and performance, the massiveness of these neural language models prove to be a challenge when transferring numerous weights and fine-tuning hundreds of millions of hyper-parameters \cite{YosinskiCBL14}. 
Furthermore, LeCun \cite{LeCun89BrainDamage} shows that only a fraction of neurons is required to produce a generalized representation for a specific task to be used for transfer learning from one data source to another.  
In light of the above observations, we hypothesize that not all representational dimensions, referred to as neurons in this paper, in the neural language model contain the same amount of linguistic information needed for solving and  transferring different levels of classification tasks.

To determine whether there are task-specific neurons in the language model that can be exploited for unsupervised transfer learning, we select the most important neurons for solving a task and generalise them across multiple data sources to a target data source.  

First, we present a \emph{single source task-specific neuron selection} (\ourProcess[Single]) algorithm by assigning the \emph{importance metrics} to neurons by examining all layers for a single data source, which was repeated via random sampling and then accumulated to ensure stability. Then, we extend the single-source algorithm for neuron selection to multiple sources by having one neuron-importance learner per each pair of disjoint subsets of data sources, which are combined by a meta-aggregator to get the \emph{meta-importance metrics} and then followed by feature selection. We call this extended method multi-source task-specific neuron selection (\ourProcess[Multi]). 
Also, a \emph{task-specific fingerprint} is introduced from those important neurons, which represents layer activation based on the percentage of the selected neurons in each layer.

To evaluate the effectiveness of our neuron selection algorithms and the task-specific fingerprints, we applied the presented methods for unsupervised transfer learning on several classification tasks such as sentiment analysis, natural language inference, and text similarity across multiple data sources and measured their performance. 
Our results show a significant improvement in those three classification tasks for unsupervised transfer learning when the target data has a similar fingerprint as the source data.

Our contributions can be summarized as follows:
\begin{itemize}
    \item [--]\textbf{Task-Specific Neuron Selection Algorithms:}  We present our method for neuron selection from the embedding space of the language representation model for single and multiple data sources by selecting the most important neurons in all the layers for a specific classification task. 
    \item [--]\textbf{State-of-the-Art Performance in Unsupervised Transfer Learning:}  The proposed method in this paper outperforms the current state-of-the-art performance in unsupervised transfer learning for sentiment analysis without fine-tuning the large and complex neural representation model architecture. Also, based on our best knowledge, there is no report on similar experiments for unsupervised transfer learning to QQP and SciTail datasets. We conclude that unsupervised transfer learning with neuron selection can lead to better performance with just a few hundred neurons that are task-interpretable.
    \item [--]\textbf{Task-Specific Fingerprints:} We demonstrate, with a set of extensive experiments, that each classification task has a distinct task-specific fingerprint, which is used for a more accurate, simpler, and interpretable base-representation of classification tasks.  
    
\end{itemize}
\section{Related Work  Literature Review}

\textbf{Language Transformers and their Explainability} has made tremendous advancement in the past year \cite{peters2018dissecting}.  
Some recent works investigate how the syntactical and semantic information is distributed through the layers of BERT specially in the fine-grain NLP tasks such as positional information, and hierarchical-information \cite{lin2019open}, part-of-speech (POS) tagging, parsing, semantic roles, and coreference resolution tasks \cite{tenney2019bert}, syntax trees \cite{hewitt2019structural}, and dependency relations \cite{coenen2019visualizing}.
Also, with analyzing the attention in different heads of the BERT layers, \cite{kovaleva2019revealing} found out that removing some heads will improve the performance of BERT for the several NLP tasks.
We extend this analysis by detecting the most generalizable neurons for a specific task from all layers of BERT. 

\textbf{Neuron Selection} With computing the neuron importance for neural machine translation and POS, it has been shown that there exists a group of neurons which learns the common linguistic knowledge \cite{bau2018identifying, dalvi2019one}. We extend these analyses to the language transformers such as BERT with computing the neuron importance in sentiment analysis, NLI, and sentence similarities. 

\textbf{Transfer learning} has seen proven success in transferring models and features to new domains in the field of image processing. However, this is an area of growth for NLP with tremendous potential especially with the recent improvement in the text semantic representation. Existing methods focus on selecting data instances \cite{remus12domainadaptationsentiment}\cite{ ruder17dataselection}, extracting domain-invariant or domain-specific features  \cite{peng-etal-2018-cross}, multi-task learning \cite{li08multidomainA}, adversarial networks \cite{liu-etal-2018-learning}, and so on. 
Also, the transferability of the ELMo model has been studied \cite{liu2019linguistic} with pre-training the transformer at fine-grain NLP tasks and evaluating the performance of each layer for another task. It is shown that some layers in the language transformer have a better transferability property than the other layers. 
In a different approach, we demonstrate that with selecting the neurons from BERT different layers and probing their performance in various classification tasks, the model achieves a new state-of-the-art in many unsupervised transfer learning settings for the classification tasks such as sentiment analysis and natural language inference.

\textbf{Meta-learning} (learning to learn) is a long-established area \cite{thrun1998learning}\cite{schmidhuber1995learning} that has recently resurfaced in popularity due to neural networks and few-shot learning \cite{ravi2016optimization}, with methods specifically focusing on learning optimization updates of hyperparameters \cite{andrychowicz2016learning} and learning weights for transferring new tasks \cite{finn2017model}. However, only a few meta-learning methods have applied transfer learning to NLP. Specifically, these NLP methods focus on an ensemble of domain-specific networks \cite{BalajiNIPS2018MetaReg}, meta-network to capture task-specific parameters \cite{chenaaai18multitask}, combine meta-predictor on various domains \cite{kiela-etal-2018-dynamic}. 
Furthermore, meta-learning in the area of stacking generalization is frequently used in transfer learning for sentiment analysis with methods that focus upon weighting features by ensembling \cite{tan-cheng-2009-improving, xia2013featureensemblesampleselection, oliverNips18realisticeval}, ensemble bootstrapping \cite{he-etal-2018-adaptive}, and weighting k-expert feedback for new domain \cite{kim-etal-2017-domain}. This leaves room for NLP methods that generalize from multiple data sources which has been studied in this paper.

\section{Methodology}

\begin{figure}[t]
\centering
\begin{subfigure}[c]{\textwidth}
    \centering
    \includegraphics[width=0.9\linewidth]{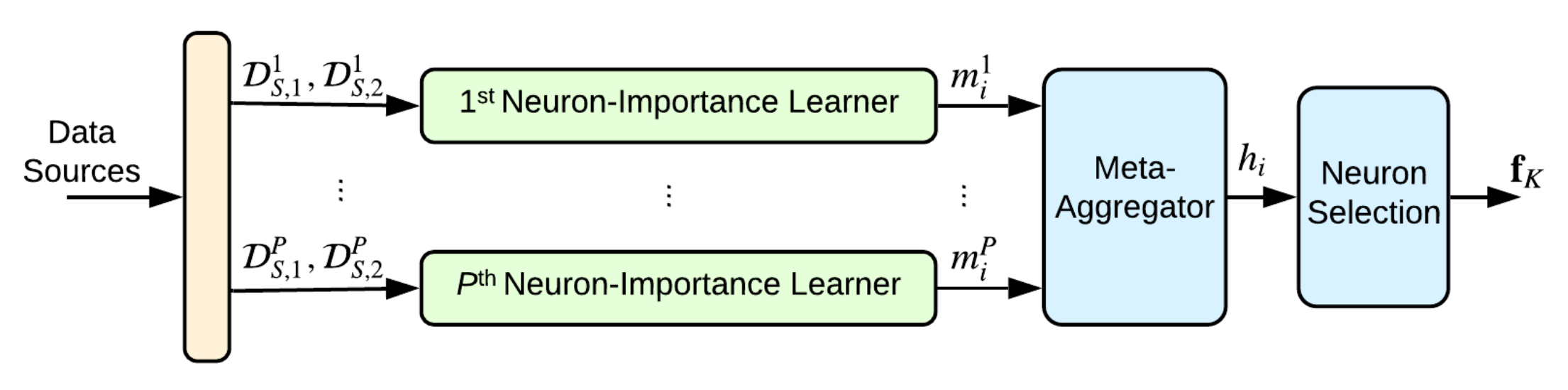}
    \caption{\ourProcess[Multi]}
    \label{fig:methodology_new_a} 
\end{subfigure}
\begin{subfigure}[c]{0.7\textwidth}
    \centering
    \includegraphics[width=1\linewidth]{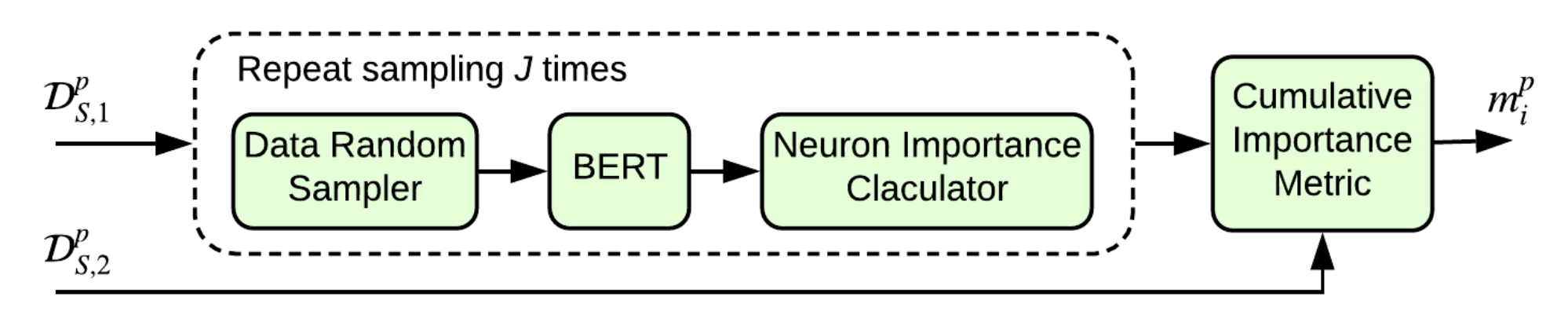}
    \caption{Internal structure of $p^{th}$ neuron-importance learner}
    \label{fig:methodology_new_b}
\end{subfigure}
\caption{Multi-source task-specific neuron selection algorithm (\ourProcess[Multi]) constructs neuron-importance learners for pairs of source dataset subsets, $(\mathcal{D}_{S,1}^p, \mathcal{D}_{S,2}^p)$.  Then, it performs meta-aggregation on the neuron importance vectors to output the meta-importance metric, $h_i$, and finally, selects the most important neurons for the task, $f_K$. Each learner computes the neuron-importance metric, $m_i^p$, for each neuron that is accumulated from iterative data random sampling.}
\label{figmethodology}
\end{figure}

To establish whether a neuron is significant for solving a specific text classification task, we present a methodology of obtaining the important neurons by pruning the representation space of BERT. The source and target datasets are called $\mathcal{D}_S$ and $\mathcal{D}_T$, respectively. 
Let $\mathcal{D} = (\textbf{X}, \textbf{y})$ be each dataset composed of texts, $\textbf{X}$,  and labels, $\textbf{y}$.
Let $\textbf{f} = [f_0, ..., f_{N-1}]$ denote the neurons of the language model. 
A general schematic of task-specific neuron selection is presented in Figure \ref{figmethodology}, which described in detail in the following sections.

\subsection{Neuron Selection for Single Source Transfer Learning}
We perform \ourProcess[Single] to select the most important neurons for the task-at-hand with only a single data source.  To ensure the stability of the importance metric, we repeatedly sample random slices of the dataset and perform feature importance on those samples. This sampling approach helps to prevent overfitting to the data source, which results in more generalizable features for transfer learning.
The \ourProcess[Single] algorithm consists of two main steps:  \emph{i}) computing the neurons' importance metric for random samples of the source distribution, and aggregating the importance metrics; and \emph{ii}) selecting the task-specific neurons from all layers of the language transformer. 

\paragraph{Neuron-Importance Learner for \ourProcess[Single]} Let us consider a neuron importance generator function $q$ that computes the importance metric for a neuron towards a specific classification task and its dataset $\mathcal{D}_S$, such that, $q_i = q(f_i, \mathcal{D}_S)$ is the importance metric of $f_i$. Then with $J$ iterations, randomly select $\beta\%$ of the source dataset $\mathcal{D}_S=(\textbf{X}_S, \textbf{y}_S)$, such that the importance metric of the $i^{th}$ neuron $f_i$ for each of the $j^{th}$ sample dataset $\mathcal{D}_S^j=(\textbf{X}_S^{j}, \textbf{y}_S^{j})$ is computed as  $q_i^{j}=q(f_i, \mathcal{D}_S^j)$. 
The cumulative importance metric is obtained for each neuron $f_i$ is:
\begin{equation}
    \textit{m}_i =\alpha E_j(q_i^{j}) + (1-\alpha) E_j(q_i^{j})/(\sigma_j(q_i^{j})+\epsilon),
\end{equation}
where $\epsilon$ is a small real number, and $E_j(.)$ and $\sigma_j(.)$ indicate the expectation and standard deviation functions with respect to $j$. In this paper, $q_i^j$s are computed using a random forest classifier applied on the outputs of all layers of the BERT language model.

\paragraph{Task-Specific Neuron Selection} Finally, the neurons with the maximum importance metrics, $m_i$, are used toward selecting $K$ task-specific neurons, $\mathbf{f}_K=[f_{i_1}, ..., f_{i_{K}}]$. 

The parameters $\alpha$ and $\beta$ are hyper-parameters set between $[0, 1]$; for \ourProcess[Single] $\alpha$ is set to $1$.  

\subsection{Neuron Selection for Multiple Source Transfer Learning}
We extend our \ourProcess[Single] to multi-source task-specific neuron selection (\ourProcess[Multi]) under the assumption that multiple data sources are more generalizable and prevent overfitting or negative transfer by any single data source. \ourProcess[Multi] uses several neuron-importance learners to transfer between two disjoint subsets of the source datasets (one for training, and one for hyperparameter tuning) in order to select neurons from different layers of the language transformer. The \ourProcess[Multi] algorithm is described in the following steps. The $M$ source datasets and the target dataset are called $\mathcal{D}_{m}$, where $m \in \{1, 2, \cdots, M\}$, and $\mathcal{D}_{T}$, respectively. 

\paragraph{Multiple Neuron-Importance Learners} Consider $P$ pairs of disjoint subsets of source datasets, $(\mathcal{D}_{S,1}^p, \mathcal{D}_{S,2}^p)$ such that $\mathcal{D}_{S,1}^p, \mathcal{D}_{S,2}^p\subset\{D_1, \cdots, D_M\}$ and $p\in\{1, \cdots, P\}$. 
The \ourProcess[Multi] consists of $P$ neuron-importance learners, described in \ourProcess[Single]. The $p^{th}$ neuron-importance learner computes the neuron-importance metric $m_i^p$ for each neuron $f_i$ using $\mathcal{D}_{S,1}^p$ datasets as the source data and randomly sample $\gamma\%$ of $\mathcal{D}_{S,2}^p$ to optimize the hyperparameter $\alpha$  based on the classification performance.

Next, we define the \emph{meta-importance metric} $h_i^p$ for neuron $f_i$, which is calculated based on the ranking of $m_i^p$ in all $m_l^p$s for $l \in \{1, \cdots, N\}$ such that $h_j^p > h_i^p$ if $f_j$ is more important than $f_i$, or in other words $m_j^p > m_i^p$. 
In out experiment, we simply calculate $ h_i^p= (2N-r_i^p)/N$, where $r_i^p$ is the rank of $m_i^p$ in all $m_l^p$s for $l \in \{1, \cdots, N\}$.

\paragraph{Meta Aggregation and Neuron Selection} The $P$ meta-importance metrics are aggregated as $h_i=\sum_p h_i^p$, for each neuron $f_i, i\in\{1, \cdots, N\}$. Then, the neurons with the maximal $h_i$s are selected as the task-specific neurons $\mathbf{f}_K$. 

The hyperparameters of this \ourProcess[Multi] algorithm are the number of neuron-importance learners ($P$), the number of selected neurons ($K$), and $\gamma$. 
\subsection{Final Training and Transferring to Target Dataset}
In order to transfer the classifier trained on the source training dataset to predict the target test dataset, we perform the following steps. First, the language model transformer is applied to the text of the source data to acquire all the neurons in all the layers. Then, those language model representations of the source texts are reduced using the $K$ selected neurons, by either our \ourProcess[Single] or our \ourProcess[Multi] and further trained by a logistic regression classifier on the labels for that specific source training task. 
The trained classifier is applied to the target data using just the $K$ selected neurons to get the predicted labels of the target data.  

\section{Experimental Results and Discussions}

\begin{table}[t]
  \caption{Dataset specifications for each NLP classification tasks in our experiments \newline}
  \label{dataset_size}
  \small
  \centering
  \begin{tabular}{p{8cm}rr}
    \toprule
    Dataset Name    & Training     & Transferring \\
    \midrule
    Amazon product reviews \cite{blitzer2007biographies} & $2,000$ each*  & $2,000$ each* \\ \hspace{4mm}(Books, DVD, Electronics, and Kitchen appliance) & & \\
    Movie review (MR) \cite{maas-EtAl:2011:ACL-HLT2011}     & $25,000$ &    \\
    SST-Binary \cite{socher2013recursive}     & $6,920$   \\
    \midrule
    SciTail \cite{khot2018scitail} & $23,596$   &   $2,126$ \\
    Stanford natural language inference (SNLI) \cite{snli:emnlp2015}  & $366,180^{\dagger}$ &      \\
    Multi-genre natural language inference (MNLI) \cite{williams2018broad}    & $261,799^{\ddagger}$ &  \\
    Recognizing textual entailment (RTE) \cite{wang2018glue} & $2,490$       &  \\
    \midrule
    Quora question pairs (QQP) \cite{wang2018glue}     & $363,177$ & $40,371^{\diamond}$    \\
    Microsoft research paraphrase corpus (MRPC) \cite{dolan2005automatically}     & $3,917$       &  \\
    \bottomrule
    \multicolumn{3}{l}{\footnotesize{* 2000 labelled examples are used as either source data or the target data. }}\\
    \multicolumn{3}{l}{\footnotesize{$^{\dagger}$ Contradiction pairs are removed from the dataset to make it analogous with SciTail and QQP. }}\\
    \multicolumn{3}{l}{\footnotesize{$^{\ddagger}$ Similarly, the classes are limited to the entailment and independent semantic}.}\\
    \multicolumn{3}{l}{\footnotesize{$^{\diamond}$ Development samples of the GLUE benchmark have been used as the testing} set.}\\
  \end{tabular}
\end{table}

To determine the effectiveness of the selected neurons by \ourProcess[Single] and \ourProcess[Multi] algorithms, we perform a set of unsupervised transfer learning experiments for the NLP classification tasks. These experiments do not utilize any data, including the validation data, from the target task:  neither in the task-specific neuron selection, the hyper-parameter tuning of algorithms, nor the training of the final classifier.  
\subsection{Materials: Task Descriptions and Datasets}

Table \ref{dataset_size} includes the different datasets for sentiment analysis, natural language inference, and text similarity (often known as paraphrasing) in the experiments.

\textbf{Sentiment Analysis (SA)} classifies  a text to its sentiment polarity. The \textit{Amazon product reviews} dataset is widely used for the multi-domain (MD) transfer learning for sentiment analysis by modifying the rating of $1$ - $5$ stars into binary sentiments \cite{blitzer2007biographies}. In contrast to the existing literature for transfer learning of
sentiment analysis, the unlabeled target data have not been exploited for the neuron selection nor training the classifier in our experiments. The \emph{movie review} (MR) corpus from the large IMDB movie reviews contains binary sentiments for classifications \cite{maas-EtAl:2011:ACL-HLT2011}. The \textit{SST-Binary} dataset is an adapted version of the Stanford sentiment treebank dataset \cite{socher2013recursive} with details found in \cite{barnes2017assessing}. 

\textbf{Natural Language Inference (NLI)} determines whether if the relations between the premise and the hypothesis texts are either entailment, contradiction, or semantic independence. \textit{SciTail} is an entailment dataset created from multiple-choice science exam and web sentences \cite{khot2018scitail} 
. The \textit{Stanford natural language inference} (SNLI) \cite{snli:emnlp2015} corpus consists of human-written labeled sentence pairs. 
The \textit{Multi-genre natural language inference} (MNLI) corpus covers a wide range of genres of formal and informal text \cite{williams2018broad} that includes a cross-genre generalization evaluation. 
The \textit{recognizing textual entailment} (RTE) datasets are collected from a series of annual RTE challenge that includes various online news sources \cite{wang2018glue}.   

\textbf{Text Similarity and Paraphrasing} determines whether two texts, such as pairs of sentences or questions, are semantically similar or paraphrase of each other. The \textit{Quora question pairs} (QQP) dataset is an unbalanced text similarity dataset for identifying duplicate questions. 
The \textit{Microsoft research paraphrase corpus (MRPC)} is a paraphrase identification dataset extracted from the news sources on the web along with human annotation \cite{dolan2005automatically}.

\subsection{Implementation Details}
The neurons of the classification token of large BERT transformer in all of its 24 layers are selected as $\textbf{f}$ resulting in $N$ as $24,576$. To ensure high-quality sub-samples of the source distributions is selected in each iteration, $\beta$ was set to $0.7$ and $0.1$ for the small and large dataset in \ourProcess[Single] and \ourProcess[Multi], respectively. Furthermore, $\alpha$ is set to $1$ for \ourProcess[Single], and optimized as described in \ourProcess[Multi]. The default value for $J$ and $\gamma$, $\epsilon$ are $100$, $10$, and $10^{-6}$ respectively. The number of neuron-importance learners in the \ourProcess[Multi] is determined by the number of the pairs of disjoint subsets in $\mathcal{D}$ as $\mathcal{D}_{S,1}^p$ and $\mathcal{D}_{S,2}^p$ for $p\in\{1, \cdots, M\}$. Finally, the random forest classifier with $1000$ trees with a depth of $200$ was used to calculate the importance metrics of neurons, $q(\cdot)$, in \ourProcess[Single]. The presented results for unsupervised transfer learning are the average of $5$ independent runs of algorithms. Micro accuracy is used for reporting the performance during experiments. An Intel Xeon Platinum 8175M series processor and a V100 Tensor Core GPU are used for the simulation.

\subsection{Task-Specific Fingerprints}
\begin{figure}[t!]
\centering
\includegraphics[width=0.9\textwidth]{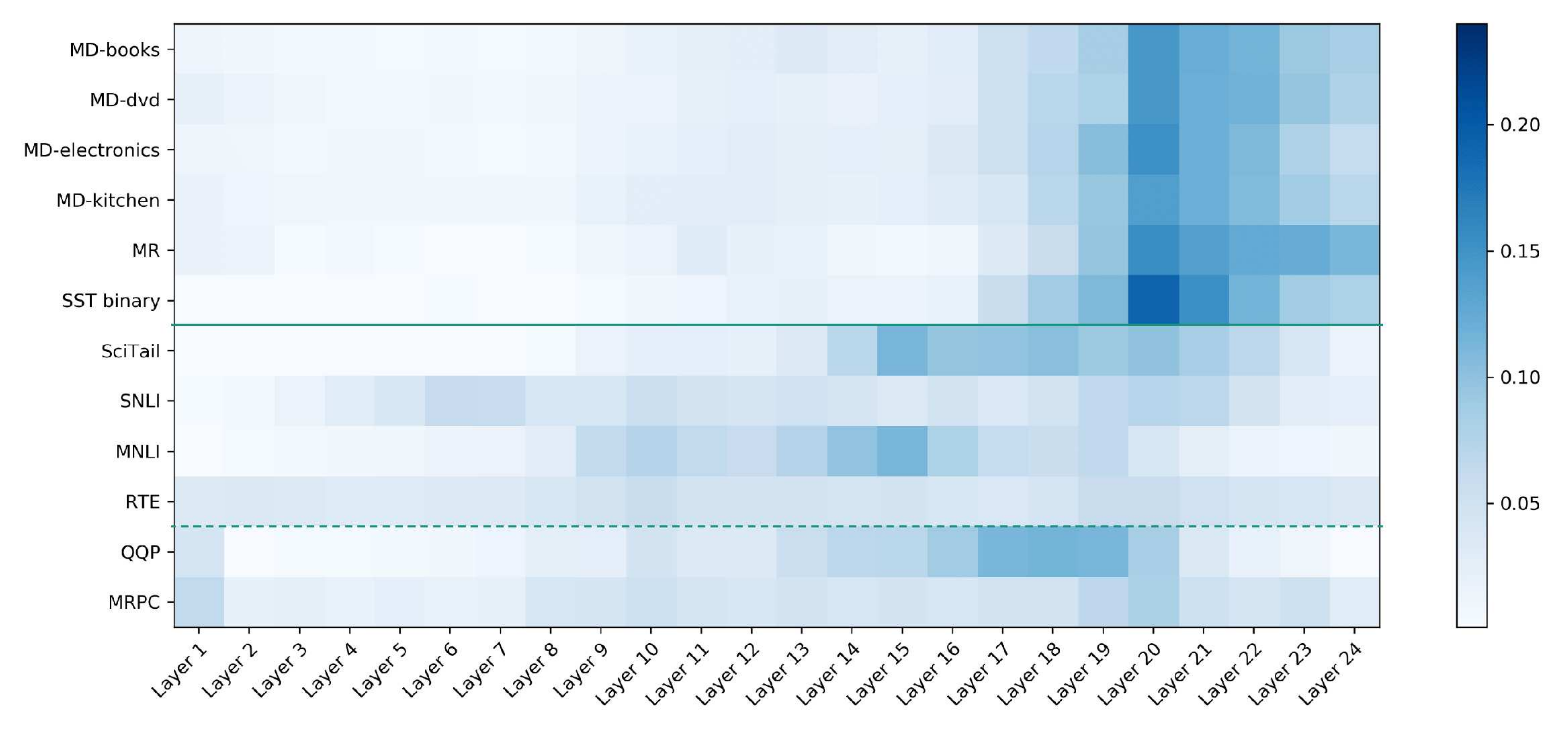}
\caption{Task-specific fingerprints of the BERT language model for different NLP classification tasks. The color of each cell shows the probability of selecting the neurons in a layer for a task's data source.}
\label{figdatasetsfingerprints}
\end{figure}
The task-specific fingerprint is defined as the percentage of the task-specific neurons in each layer of the language model. Fig.~\ref{figdatasetsfingerprints}  displays the task-specific fingerprints for various classification tasks and their data sources for the BERT language model. 
It can be observed from the task-specific fingerprints that different tasks have specific regions of activation in the layers of the language model due to their required specific syntactical and semantic information. For sentiment analysis, the last six layers are the most activated layers for different domains of MD, MR, and SST-binary datasets. This observation is inconsistent with existing literature \cite{peters2018dissecting} that suggests lower layers captures short-range dependencies while upper layers capture long-range dependencies.

However, contrary to the general hypothesis 
that the task-dependent neurons are typically attached to the last upper layer of the neural architecture, the fingerprints for the SciTail, MNLI, and QQP datasets are most significant in the middle layers. This could be due to the middle layers carrying information of the syntactic structure of sentences for NLI and text similarity tasks \cite{hewittstructural}. Lastly, the task-specific neurons are distributed over almost all layers of SNLI, RTE, and MRPC datasets due to their small or human-written sentences. 

The task-specific fingerprints can be used for interpretations as well as giving hints for combining different NLP tasks and corresponding datasets for achieving better performance for transfer learning. 

\subsection{Unsupervised Transfer Learning with Task-Specific Neurons}
To evaluate the \ourProcess[Single] and \ourProcess[Multi] algorithms for training more generalized models, extensive unsupervised transfer learning experiments were conducted on a wide variety of tasks. We also interpret results with the task-specific fingerprints of the source and target datasets.  

\begin{table}[t]
\centering
\caption{Single source unsupervised transfer learning on the multi-domain sentiment analysis dataset. \newline  }
\label{Tabel:SingleSourceSentiment}
\small
\begin{tabular}{lm{0.047\textwidth}m{0.047\textwidth}m{0.047\textwidth}m{0.047\textwidth}m{0.01\textwidth}m{0.05\textwidth}m{0.005\textwidth}m{0.047\textwidth}m{0.047\textwidth}m{0.047\textwidth}m{0.047\textwidth}m{0.047\textwidth}}
\toprule
\multirow{2}{*}{S ---> T} & \multirow{1}{*}{CDB} & \multirow{1}{*}{CMD} & \multirow{1}{*}{PBLM} & \multirow{1}{*}{PE} & & BERT  && \multicolumn{5}{l}{\hspace{15mm}SingleTSNS (ours)} \\
\cmidrule{6-7} \cmidrule{9-13}
 &\cite{ruder17learningselect}&\cite{ZellingerGLNS17cmd}&\cite{ziser-reichart-2018-pivot}&\cite{barnes-etal-2018-projecting} &$K$ & 1024 &  & 100  & 300  & 500 & 700 & 1024 \\
\midrule
D ---> B    &74.2   &79.5&82.5&77.3&&\textbf{82.6}&&80.5&81.6&81.2&81.3&80.7\\
E ---> B    &74.1   &74.4&71.4&71.2&&62.6&&75.5&\textbf{77.9}&73.8&71.3&71.0\\
K ---> B    &75.5   &75.6&74.2&73.0&&75.1&&73.9&77.7&77.7&78.5&\textbf{79.3}\\
B ---> D    &64.8   &80.5&84.2&81.1&&83.0&&80.1&83.7&83.5&84.0&\textbf{84.4}\\
E ---> D    &\textbf{79.3}   &76.3&75.0&74.5&&73.6&&78.0&77.7&78.3&77.7&77.0\\
K ---> D    &76.1   &77.5&\textbf{79.8}&76.3&&71.0&&69.8&73.2&77.5&78.5&79.3\\
B ---> E    &65.1   &78.7&77.6&76.8&&75.9&&80.6&80.8&\textbf{81.8}&78.4&76.5\\
D ---> E    &\textbf{83.3}&79.7&79.6&78.1&&81.4&&70.7&80.1&82.4&81.8&80.0\\
K ---> E    &78.7   &85.4&\textbf{87.1}&84.0&&84.1&&83.6&85.1&86.4&85.9&85.2\\
B ---> K    &79.9   &81.3&82.5&80.1&&75.2&&82.2&\textbf{85.2}&85.0&84.5&83.2\\
D ---> K    &\textbf{85.2}&83.0&83.2&80.3&&84.4&&75.7&83.2&84.6&83.8&83.5\\
E ---> K    &84.5   &86.0&\textbf{87.8}&84.6&&75.1&&84.4&86.1&86.1&87.2&87.1\\
\midrule
Average&76.7&79.9&80.4&78.2&&77.4&&77.9&81.0&\textbf{81.5}&81.1&80.6\\
\bottomrule
\multicolumn{10}{l}{\footnotesize{B, D, E, K denote Book, DVD, Electronics, and Kitchen Appliance, respectively.}}
\end{tabular}
\end{table}
\paragraph{Single Source Transfer learning}
Table \ref{Tabel:SingleSourceSentiment} shows that for unsupervised transfer learning applied to sentiment analysis. The selected neurons by \ourProcess[Single] outperform the previous state-of-the-art in 5 out of 12 pairs of domains for the MD dataset. It also outperforms the accuracy of the average of all $12$ transferring tasks by $1.1\%$ when compared to the literature and $4.1\%$ when compared to the last layer BERT.
This improvement indicates that the selected neurons carry more generalized information regarding the sentiment analysis task. Also, based on the fingerprints of the sentiment analysis datasets in Fig. \ref{figdatasetsfingerprints}, results show that including $20^{th}$-$22^{th}$ layers lead to a better sentiment prediction. Additionally, tasks or domains that have similar patterns in the activated layers of their task-specific fingerprints are the best candidates for transferring among them. For example, the best domain for transferring to Books is DVD and vice versa. Similarly, the best domain for transferring to Kitchen is Electronics and vice versa. 

\begin{table}[]
\centering
\caption{Single source unsupervised transfer learning to the QQP and SciTail. \newline}
\small
\label{Tabel:SingleSourceNLI}
\begin{tabular}{m{0.055\textwidth}m{0.055\textwidth}m{0.045\textwidth}m{0.002\textwidth}m{0.045\textwidth}m{0.045\textwidth}m{0.045\textwidth}m{0.055\textwidth}m{0.045\textwidth}m{0.002\textwidth}m{0.045\textwidth}m{0.045\textwidth}m{0.045\textwidth}}
\toprule
\multirow{2}{*}{Source} & \multirow{2}{*}{Target} &  BERT & \multirow{2}{*}{} & \multicolumn{3}{l}{\hspace{1mm}SingleTSNS (ours)} &  \multirow{2}{*}{Target} &  BERT  &\multirow{2}{*}{} & \multicolumn{3}{l}{\hspace{1 mm}SingleTSNS (ours)} \\
\cmidrule{3-3}\cmidrule{5-7} \cmidrule{9-9} \cmidrule{11-13}
 & & 1024 & & 100 & 500 & 1024 & & 1024 &  & 100  & 500 & 1024 \\
\midrule
QQP     &   SciTail &66.9   && 63.1   &67.6 &\textbf{71.9}   &QQP    &  ---     &&---   &---    &--- \\
SciTail &   SciTail &---    && ---    &---  &---    &QQP    &  62.3    &&63.3  &64.9   &\textbf{63.7}\\
MNLI    &   SciTail &62.7   && 66.0   &72.9 &\textbf{73.2}   &QQP    &  63.0    &&64.6  &64.4   &\textbf{66.0}\\
SNLI    &   SciTail &52.5   && 60.3   &\textbf{63.8} &55.7   &QQP    &  52.0    &&58.1  &\textbf{60.8}   &60.7\\
MRPC    &   SciTail &55.7   && \textbf{62.9}   &56.7 &57.2   &QQP    &  61.2    &&55.0  &57.0   &\textbf{62.3}\\
RTE     &   SciTail &59.4   && 62.0   &\textbf{62.8} &59.5   &QQP    &  45.7   &&\textbf{47.6}  &40.17   &43.61\\
\bottomrule
\end{tabular}
\end{table}

Unsupervised transfer learning was applied to the SciTail and QQP datasets, which are NLI and text similarity tasks respectively.

Fig. \ref{figdatasetsfingerprints} shows that MNLI, SciTail, and QQP have an effective representation space at the middle-upper layers, i.e., $14^{th}$-$20^{th}$ layers of the BERT transformer. As a result in Table \ref{Tabel:SingleSourceNLI}, they are the best datasets for transfer learning amongst them. 
Furthermore, when comparing to selecting neurons in the last layer of BERT as a baseline, the accuracy increases significantly. Based on our best knowledge, there is not any other reference for completely unsupervised transfer learning to QQP and SciTail datasets without fine-tuning the language transformer.

One observation that emerged from these experiments is that the performance of transfer learning also depends upon additional factors such as the size and the composition of datasets. For example, MNLI is a large dataset consisting of a mixture of multiple domains, thus giving MNLI an inherent advantage as a source dataset for transfer learning. Similarly, RTE and MPRC are small datasets, thus they perform poorly when transferring to a large target dataset. Interestingly, the results in the next section demonstrate that the combination of small data sources with other sources in \ourProcess[Multi] can improve the accuracy of transfer learning. 

\paragraph{Multiple Source Transfer Learning}
\begin{table}[t]
\centering
\caption{Multi-source unsupervised transfer learning for sentiment analysis. \newline  }
\small
\label{Tabel:MultiSourceSentiment}
\begin{tabular}{lm{0.01\textwidth}m{0.06\textwidth}m{0.06\textwidth}m{0.06\textwidth}m{0.01\textwidth}m{0.05\textwidth}m{0.005\textwidth}m{0.047\textwidth}m{0.047\textwidth}m{0.047\textwidth}m{0.047\textwidth}}
\toprule
\multirow{2}{*}{S ---> T} & & \multirow{1}{*}{SDAM} & \multirow{1}{*}{MDAN} & \multirow{1}{*}{MTLE} & & BERT  && \multicolumn{4}{l}{\hspace{7mm}MultiTSNS (ours)} \\
\cmidrule{6-7} \cmidrule{9-12}
 & &\cite{wu-huang-2016-sentiment}&\cite{zhao2018adversarial}&\cite{zhang2018multi} &$K$ & 1024 &  & 200  & 500 & 700 & 1024 \\
\midrule
DEK ---> B&&78.3&78.6&---&&80.6&&82.2&\textbf{83.0}&81.7&80.6\\
BEK ---> D&&79.1&80.7&---&&83.5&&83.8&84.8&\textbf{85.3}&83.5\\
BDK ---> E&&84.2&85.3&---&&83.4&&85.0&86.1&\textbf{86.8}&83.6\\
BDE ---> K&&86.3&86.3&86.7&&80.9&&\textbf{87.9}&86.5&86.2&80.9\\
\midrule
Average&&82.0&82.7&---&&82.1&&84.7&\textbf{85.1}&85.0&82.2\\
\bottomrule
\multicolumn{12}{l}{\footnotesize{MD sentiment dataset, where ABC-->D denotes to transferring from domains A, B, and C to D. } }\\

\end{tabular}
\end{table}
Tables \ref{Tabel:MultiSourceSentiment} and \ref{Tabel:MultiSourceNLI} show the sentiment analysis and NLI results, respectively, for unsupervised transfer learning with the selected neurons by \ourProcess[Multi]. 
As it can be seen from the results in Table \ref{Tabel:MultiSourceSentiment}, \ourProcess[Multi] outperforms the state-of-the-art for each of the transferring scenarios as well as for the overall average. Furthermore, selecting 500 neurons from all layers achieves a $3.0\%$ average improvement over using the 1024 last-layer neurons in BERT. 

\begin{table}[t]
\centering
\caption{Multi-source unsupervised transfer learning to SciTail. The underlines and bold values determine the best model for each combination of sources and the best sources for each model, respectively. \newline  }
\label{Tabel:MultiSourceNLI}
\small
\begin{tabular}{m{0.250\textwidth}m{0.055\textwidth}m{0.002\textwidth}m{0.01\textwidth}m{0.045\textwidth}m{0.002\textwidth}m{0.045\textwidth}m{0.045\textwidth}m{0.045\textwidth}m{0.001\textwidth}m{0.045\textwidth}m{0.045\textwidth}m{0.045\textwidth}}
\toprule
\multirow{3}{*}{Source} & \multirow{3}{*}{Target}& \multirow{3}{*}{} & \multirow{3}{*}{} &  BERT & \multirow{3}{*}{} & \multicolumn{7}{l}{\hspace{19mm}MultiTSNS (ours)}\\
\cmidrule{4-5}\cmidrule{7-13}
 & && L &2E5 & & \multicolumn{3}{l}{\hspace{10mm}2E5} & & 2E4  & 1E5 & 3E5 \\
\cmidrule{4-5}\cmidrule{7-9} \cmidrule{11-13}
 & && K&1024 & & 100 & 500 & 1024 & & 500  & 500 & 500 \\
\midrule
QQP     &   SciTail &&&\textbf{68.5}	&&62.8	&67.6	&\underline{71.8}	&&68.5	&67.1	&67.6 \\
MNLI    &   SciTail &&&62.7	&&66.3	&72.9	&\underline{73.1}	&&62.8	&72.4	&72.9 \\
SNLI    &   SciTail &&&50.8	&&60.3	&\underline{64.3}	&55.5	&&50.8	&64.0	&63.7 \\
\midrule
QQP+RTE              &   SciTail &&&66.7	&&64.2	&70.0   &\underline{71.5}	&&70.8	&71.0	&70.5 \\
QQP+MNLI             &   SciTail &&&63.9	&&69.2	&\textbf{76.9}	&75.1	&&73.3	&\textbf{75.8}	&\underline{\textbf{77.1}} \\
QQP+MNLI+RTE         &   SciTail &&&64.0	&&\textbf{72.2}	&74.4	&\underline{75.6}	&&71.1	&73.5	&73.4 \\
QQP+MNLI+RTE+MPRC    &   SciTail &&&64.7	&&70.9	&74.8	&\underline{\textbf{75.3}}	&&\textbf{73.4}	&75.2	&74.6 \\
QQP+SNLI             &   SciTail &&&62.8	&&69.1	&\underline{70.7}	&68.4	&&67.6	&69.7	&70.6 \\
QQP+MNLI+SNLI        &   SciTail &&&63.7	&&68.0	&\underline{73.1}	&72.0	&&70.8	&72.7	&\underline{73.1} \\
\bottomrule
\end{tabular}
\end{table}
Table \ref{Tabel:MultiSourceNLI}  presents the results of transferring by \ourProcess[Multi] from different combinations of multiple source datasets to the target task of SCiTail dataset for NLI.
For a fair comparison between different combinations of source datasets for transferring to the target dataset, the maximum total number of source data samples in the MultiTSNS algorithm and the final classifier training was limited. The water-filling algorithm is used to allocate the limitation between sources \cite{Proakis01}. 
The results improved by $3.5-9.5\%$ in comparison to \ourProcess[Single], which indicates that using multiple datasets increases the coverage of the target activation layers in the task-specific fingerprint. For example, when transferring to the SciTail  dataset, including the MNLI and the RTE in source datasets results in a better coverage by their respective task-specific fingerprints, i.e. the middle-upper layers and all the layers respectively.
Lastly, when compared to the BERT baseline, which uses just the last layer, the \ourProcess[Multi] algorithm increases the performance by $3.8-13.0\%$ for different data source combinations. We also tested the sensitivity of \ourProcess[Multi] algorithm to the number of source samples by changing the maximum number of source samples from $20,000$ to $300,000$ in Table \ref{Tabel:MultiSourceNLI}. 
\section{Conclusion and Future Work}
This paper introduces a new transfer learning methodology with selecting task-specific neurons from all layers of language transformers to generalize the trained models for unsupervised transfer learning in NLP classification tasks (i.e. sentiment analysis, natural language inference, and sentence similarities).
Our results show that each task data source has a distinct task-specific fingerprint of the activation layers in the BERT language model. 
These pieces of evidence imply that neural networks can be refined for a specific task due to the semantic and syntactic knowledge that is hidden in the representational space throughout all their layers. Furthermore, our experiments on combining importance metrics of multiple subsets of datasets using a meta-aggrigator suggests a promising approach for transferring multiple data sources to a single target task dataset. Future work can extend the algorithms to other NLP tasks such as sequence labelling and to other meta-learning architectures, and the experiments to additional transfer learning scenarios such as semi-supervised training.  

\bibliographystyle{plain}


\begin{thebibliography}{10}

\bibitem{andrychowicz2016learning}
Marcin Andrychowicz, Misha Denil, Sergio Gomez, Matthew~W Hoffman, David Pfau,
  Tom Schaul, Brendan Shillingford, and Nando De~Freitas.
\newblock Learning to learn by gradient descent by gradient descent.
\newblock In {\em Advances in Neural Information Processing Systems}, pages
  3981--3989, 2016.

\bibitem{BalajiNIPS2018MetaReg}
Yogesh Balaji, Swami Sankaranarayanan, and Rama Chellappa.
\newblock Metareg: Towards domain generalization using meta-regularization.
\newblock In S.~Bengio, H.~Wallach, H.~Larochelle, K.~Grauman, N.~Cesa-Bianchi,
  and R.~Garnett, editors, {\em Advances in Neural Information Processing
  Systems 31}, pages 998--1008. Curran Associates, Inc., 2018.

\bibitem{barnes2017assessing}
Jeremy Barnes, Roman Klinger, and Sabine~Schulte im~Walde.
\newblock Assessing state-of-the-art sentiment models on state-of-the-art
  sentiment datasets.
\newblock In {\em Proceedings of the 8th Workshop on Computational Approaches
  to Subjectivity, Sentiment and Social Media Analysis}, pages 2--12, 2017.

\bibitem{barnes-etal-2018-projecting}
Jeremy Barnes, Roman Klinger, and Sabine Schulte~im Walde.
\newblock Projecting embeddings for domain adaption: Joint modeling of
  sentiment analysis in diverse domains.
\newblock In {\em Proceedings of the 27th International Conference on
  Computational Linguistics}, pages 818--830, Santa Fe, New Mexico, USA, August
  2018. Association for Computational Linguistics.

\bibitem{bau2018identifying}
Anthony Bau, Yonatan Belinkov, Hassan Sajjad, Nadir Durrani, Fahim Dalvi, and
  James Glass.
\newblock Identifying and controlling important neurons in neural machine
  translation.
\newblock In {\em International Conference on Learning Representations}, 2019.

\bibitem{blitzer2007biographies}
John Blitzer, Mark Dredze, and Fernando Pereira.
\newblock Biographies, bollywood, boom-boxes and blenders: Domain adaptation
  for sentiment classification.
\newblock In {\em Proceedings of the 45th annual meeting of the association of
  computational linguistics}, pages 440--447, 2007.

\bibitem{bottou2014machine}
L{\'e}on Bottou.
\newblock From machine learning to machine reasoning.
\newblock {\em Machine learning}, 94(2):133--149, 2014.

\bibitem{snli:emnlp2015}
Samuel~R Bowman, Gabor Angeli, Christopher Potts, and Christopher~D Manning.
\newblock A large annotated corpus for learning natural language inference.
\newblock {\em arXiv preprint arXiv:1508.05326}, 2015.

\bibitem{chenaaai18multitask}
Junkun Chen, Xipeng Qiu, Pengfei Liu, and Xuanjing Huang.
\newblock Meta multi-task learning for sequence modeling.
\newblock In {\em Thirty-Second AAAI Conference on Artificial Intelligence},
  2018.

\bibitem{coenen2019visualizing}
Andy Coenen, Emily Reif, Ann Yuan, Been Kim, Adam Pearce, Fernanda Vi{\'e}gas,
  and Martin Wattenberg.
\newblock Visualizing and measuring the geometry of bert.
\newblock {\em arXiv preprint arXiv:1906.02715}, 2019.

\bibitem{dalvi2019one}
Fahim Dalvi, Nadir Durrani, Hassan Sajjad, Yonatan Belinkov, D~Anthony Bau, and
  James Glass.
\newblock What is one grain of sand in the desert? analyzing individual neurons
  in deep nlp models.
\newblock In {\em Proceedings of the AAAI Conference on Artificial Intelligence
  (AAAI)}, 2019.

\bibitem{devlin19bert}
Jacob Devlin, Ming-Wei Chang, Kenton Lee, and Kristina Toutanova.
\newblock Bert: Pre-training of deep bidirectional transformers for language
  understanding.
\newblock {\em arXiv preprint arXiv:1810.04805}, 2018.

\bibitem{dolan2005automatically}
William~B Dolan and Chris Brockett.
\newblock Automatically constructing a corpus of sentential paraphrases.
\newblock In {\em Proceedings of the Third International Workshop on
  Paraphrasing (IWP2005)}, 2005.

\bibitem{finn2017model}
Chelsea Finn, Pieter Abbeel, and Sergey Levine.
\newblock Model-agnostic meta-learning for fast adaptation of deep networks.
\newblock In {\em Proceedings of the 34th International Conference on Machine
  Learning-Volume 70}, pages 1126--1135. JMLR. org, 2017.

\bibitem{he-etal-2018-adaptive}
Ruidan He, Wee~Sun Lee, Hwee~Tou Ng, and Daniel Dahlmeier.
\newblock Adaptive semi-supervised learning for cross-domain sentiment
  classification.
\newblock In {\em Proceedings of the 2018 Conference on Empirical Methods in
  Natural Language Processing}, pages 3467--3476, Brussels, Belgium,
  October-November 2018. Association for Computational Linguistics.

\bibitem{hewitt2019structural}
John Hewitt and Christopher~D Manning.
\newblock A structural probe for finding syntax in word representations.
\newblock In {\em Proceedings of the 2019 Conference of the North American
  Chapter of the Association for Computational Linguistics: Human Language
  Technologies, Volume 1 (Long and Short Papers)}, pages 4129--4138, 2019.

\bibitem{hewittstructural}
John Hewitt and Christopher~D Manning.
\newblock A structural probe for finding syntax in word representations.
\newblock In {\em Proceedings of the 2019 Conference of the North American
  Chapter of the Association for Computational Linguistics: Human Language
  Technologies, NAACL-HLT, Minneapolis, Minnesota, USA}, volume~2, 2019.

\bibitem{khot2018scitail}
Tushar Khot, Ashish Sabharwal, and Peter Clark.
\newblock Scitail: A textual entailment dataset from science question
  answering.
\newblock In {\em Thirty-Second AAAI Conference on Artificial Intelligence},
  2018.

\bibitem{kiela-etal-2018-dynamic}
Douwe Kiela, Changhan Wang, and Kyunghyun Cho.
\newblock Dynamic meta-embeddings for improved sentence representations.
\newblock In {\em Proceedings of the 2018 Conference on Empirical Methods in
  Natural Language Processing}, pages 1466--1477, Brussels, Belgium,
  October-November 2018. Association for Computational Linguistics.

\bibitem{kim-etal-2017-domain}
Young-Bum Kim, Karl Stratos, and Dongchan Kim.
\newblock Domain attention with an ensemble of experts.
\newblock In {\em Proceedings of the 55th Annual Meeting of the Association for
  Computational Linguistics (Volume 1: Long Papers)}, pages 643--653,
  Vancouver, Canada, July 2017. Association for Computational Linguistics.

\bibitem{kovaleva2019revealing}
Olga Kovaleva, Alexey Romanov, Anna Rogers, and Anna Rumshisky.
\newblock Revealing the dark secrets of bert.
\newblock {\em arXiv preprint arXiv:1908.08593}, 2019.

\bibitem{LeCun89BrainDamage}
Yann LeCun, John~S. Denker, and Sara~A. Solla.
\newblock Optimal brain damage.
\newblock In D.~S. Touretzky, editor, {\em Advances in Neural Information
  Processing Systems 2}, pages 598--605. Morgan-Kaufmann, 1990.

\bibitem{li08multidomainA}
Shoushan Li and Chengqing Zong.
\newblock Multi-domain sentiment classification.
\newblock In {\em Proceedings of the 46th Annual Meeting of the Association for
  Computational Linguistics on Human Language Technologies: Short Papers},
  HLT-Short '08, pages 257--260, Stroudsburg, PA, USA, 2008. Association for
  Computational Linguistics.

\bibitem{lin2019open}
Yongjie Lin, Yi~Chern Tan, and Robert Frank.
\newblock Open sesame: Getting inside bert's linguistic knowledge.
\newblock {\em arXiv preprint arXiv:1906.01698}, 2019.

\bibitem{liu2019linguistic}
Nelson~F Liu, Matt Gardner, Yonatan Belinkov, Matthew Peters, and Noah~A Smith.
\newblock Linguistic knowledge and transferability of contextual
  representations.
\newblock {\em arXiv preprint arXiv:1903.08855}, 2019.

\bibitem{liu-etal-2018-learning}
Qi~Liu, Yue Zhang, and Jiangming Liu.
\newblock Learning domain representation for multi-domain sentiment
  classification.
\newblock In {\em Proceedings of the 2018 Conference of the North {A}merican
  Chapter of the Association for Computational Linguistics: Human Language
  Technologies, Volume 1 (Long Papers)}, pages 541--550, New Orleans,
  Louisiana, June 2018. Association for Computational Linguistics.

\bibitem{maas-EtAl:2011:ACL-HLT2011}
Andrew~L. Maas, Raymond~E. Daly, Peter~T. Pham, Dan Huang, Andrew~Y. Ng, and
  Christopher Potts.
\newblock Learning word vectors for sentiment analysis.
\newblock In {\em Proceedings of the 49th Annual Meeting of the Association for
  Computational Linguistics: Human Language Technologies}, pages 142--150,
  Portland, Oregon, USA, June 2011. Association for Computational Linguistics.

\bibitem{oliverNips18realisticeval}
Avital Oliver, Augustus Odena, Colin~A Raffel, Ekin~Dogus Cubuk, and Ian
  Goodfellow.
\newblock Realistic evaluation of deep semi-supervised learning algorithms.
\newblock In S.~Bengio, H.~Wallach, H.~Larochelle, K.~Grauman, N.~Cesa-Bianchi,
  and R.~Garnett, editors, {\em Advances in Neural Information Processing
  Systems 31}, pages 3235--3246. Curran Associates, Inc., 2018.

\bibitem{peng-etal-2018-cross}
Minlong Peng, Qi~Zhang, Yu-gang Jiang, and Xuanjing Huang.
\newblock Cross-domain sentiment classification with target domain specific
  information.
\newblock In {\em Proceedings of the 56th Annual Meeting of the Association for
  Computational Linguistics (Volume 1: Long Papers)}, pages 2505--2513,
  Melbourne, Australia, July 2018. Association for Computational Linguistics.

\bibitem{peters-etal-2018-deep}
Matthew Peters, Mark Neumann, Mohit Iyyer, Matt Gardner, Christopher Clark,
  Kenton Lee, and Luke Zettlemoyer.
\newblock Deep contextualized word representations.
\newblock In {\em Proceedings of the 2018 Conference of the North {A}merican
  Chapter of the Association for Computational Linguistics: Human Language
  Technologies, Volume 1 (Long Papers)}, pages 2227--2237, New Orleans,
  Louisiana, June 2018. Association for Computational Linguistics.

\bibitem{peters2018dissecting}
Matthew~E. Peters, Mark Neumann, Luke Zettlemoyer, and Wen{-}tau Yih.
\newblock Dissecting contextual word embeddings: Architecture and
  representation.
\newblock {\em CoRR}, abs/1808.08949, 2018.

\bibitem{Proakis01}
John~G. Proakis.
\newblock {\em Digital Communication Systems}.
\newblock McGraw Hill, 2001.

\bibitem{ravi2016optimization}
Sachin Ravi and Hugo Larochelle.
\newblock Optimization as a model for few-shot learning.
\newblock In {\em 5th International Conference on Learning Representations,
  {ICLR} 2017, Toulon, France, April 24-26, 2017, Conference Track
  Proceedings}, 2017.

\bibitem{remus12domainadaptationsentiment}
R.~{Remus}.
\newblock Domain adaptation using domain similarity- and domain
  complexity-based instance selection for cross-domain sentiment analysis.
\newblock In {\em 2012 IEEE 12th International Conference on Data Mining
  Workshops}, pages 717--723, Dec 2012.

\bibitem{ruder17dataselection}
Sebastian Ruder, Parsa Ghaffari, and John~G. Breslin.
\newblock Data selection strategies for multi-domain sentiment analysis.
\newblock {\em CoRR}, abs/1702.02426, 2017.

\bibitem{ruder17learningselect}
Sebastian Ruder and Barbara Plank.
\newblock Learning to select data for transfer learning with {B}ayesian
  optimization.
\newblock In {\em Proceedings of the 2017 Conference on Empirical Methods in
  Natural Language Processing}, pages 372--382, Copenhagen, Denmark, September
  2017. Association for Computational Linguistics.

\bibitem{schmidhuber1995learning}
Jürgen Schmidhuber.
\newblock On learning how to learn learning strategies.
\newblock Technical report, TUM – Technische Universität München, 1995.

\bibitem{socher2013recursive}
Richard Socher, Alex Perelygin, Jean Wu, Jason Chuang, Christopher~D Manning,
  Andrew Ng, and Christopher Potts.
\newblock Recursive deep models for semantic compositionality over a sentiment
  treebank.
\newblock In {\em Proceedings of the 2013 conference on empirical methods in
  natural language processing}, pages 1631--1642, 2013.

\bibitem{tan-cheng-2009-improving}
Songbo Tan and Xueqi Cheng.
\newblock Improving {SCL} model for sentiment-transfer learning.
\newblock In {\em Proceedings of Human Language Technologies: The 2009 Annual
  Conference of the North {A}merican Chapter of the Association for
  Computational Linguistics, Companion Volume: Short Papers}, pages 181--184,
  Boulder, Colorado, June 2009. Association for Computational Linguistics.

\bibitem{tenney2019bert}
Ian Tenney, Dipanjan Das, and Ellie Pavlick.
\newblock Bert rediscovers the classical nlp pipeline.
\newblock {\em arXiv preprint arXiv:1905.05950}, 2019.

\bibitem{thrun1998learning}
Sebastian Thrun and Lorien Pratt.
\newblock Learning to learn: Introduction and overview.
\newblock In {\em Learning to learn}, pages 3--17. Springer, 1998.

\bibitem{wang2018glue}
Alex Wang, Amanpreet Singh, Julian Michael, Felix Hill, Omer Levy, and Samuel~R
  Bowman.
\newblock Glue: A multi-task benchmark and analysis platform for natural
  language understanding.
\newblock {\em EMNLP 2018}, page 353, 2018.

\bibitem{williams2018broad}
Adina Williams, Nikita Nangia, and Samuel Bowman.
\newblock A broad-coverage challenge corpus for sentence understanding through
  inference.
\newblock In {\em Proceedings of the 2018 Conference of the North American
  Chapter of the Association for Computational Linguistics: Human Language
  Technologies, Volume 1 (Long Papers)}, pages 1112--1122, 2018.

\bibitem{wu-huang-2016-sentiment}
Fangzhao Wu and Yongfeng Huang.
\newblock Sentiment domain adaptation with multiple sources.
\newblock In {\em Proceedings of the 54th Annual Meeting of the Association for
  Computational Linguistics (Volume 1: Long Papers)}, pages 301--310, Berlin,
  Germany, August 2016. Association for Computational Linguistics.

\bibitem{xia2013featureensemblesampleselection}
Rui Xia, Chengqing Zong, Xuelei Hu, and Erik Cambria.
\newblock Feature ensemble plus sample selection: Domain adaptation for
  sentiment classification.
\newblock {\em IEEE Intelligent Systems}, 28(3):10--18, May 2013.

\bibitem{YosinskiCBL14}
Jason Yosinski, Jeff Clune, Yoshua Bengio, and Hod Lipson.
\newblock How transferable are features in deep neural networks?
\newblock In Z.~Ghahramani, M.~Welling, C.~Cortes, N.~D. Lawrence, and K.~Q.
  Weinberger, editors, {\em Advances in Neural Information Processing Systems
  27}, pages 3320--3328. Curran Associates, Inc., 2014.

\bibitem{ZellingerGLNS17cmd}
Werner Zellinger, Thomas Grubinger, Edwin Lughofer, Thomas Natschl{\"{a}}ger,
  and Susanne Saminger{-}Platz.
\newblock Central moment discrepancy {(CMD)} for domain-invariant
  representation learning.
\newblock In {\em {ICLR}}. OpenReview.net, 2017.

\bibitem{zhang2018multi}
Honglun Zhang, Liqiang Xiao, Wenqing Chen, Yongkun Wang, and Yaohui Jin.
\newblock Multi-task label embedding for text classification.
\newblock In {\em Proceedings of the 2018 Conference on Empirical Methods in
  Natural Language Processing}, pages 4545--4553, 2018.

\bibitem{zhao2018adversarial}
Han Zhao, Shanghang Zhang, Guanhang Wu, Jos{\'e}~MF Moura, Joao~P Costeira, and
  Geoffrey~J Gordon.
\newblock Adversarial multiple source domain adaptation.
\newblock In {\em Advances in Neural Information Processing Systems}, pages
  8559--8570, 2018.

\bibitem{ziser-reichart-2018-pivot}
Yftah Ziser and Roi Reichart.
\newblock Pivot based language modeling for improved neural domain adaptation.
\newblock In {\em Proceedings of the 2018 Conference of the North {A}merican
  Chapter of the Association for Computational Linguistics: Human Language
  Technologies, Volume 1 (Long Papers)}, pages 1241--1251, New Orleans,
  Louisiana, June 2018. Association for Computational Linguistics.

\end{thebibliography}

\end{document}